\documentclass[letterpaper, 10 pt, conference]{ieeeconf}  % Comment this line out if you need a4paper

\IEEEoverridecommandlockouts                              % This command is only needed if 
\usepackage{graphics} % for pdf, bitmapped graphics files
\usepackage{float}
\usepackage{booktabs}% for the [H] float placement specifier
\usepackage{array} % for the p{} column type with custom alignment
\usepackage{soul, color}

\makeatletter
\let\NAT@parse\undefined
\makeatother
\usepackage[numbers,sort&compress]{natbib} % provides \citep, \citet used with IEEEtranS bibliography style

\usepackage[most]{tcolorbox}
\tcbuselibrary{listings, skins, breakable}
\usepackage{xcolor}
\usepackage{caption} % for \captionof, used to place Fig. 1 in the title block

\newtcolorbox{promptbox}[1]{%
  colback=gray!5,          % Background color
  colframe=purple!75!black, % Border and title bar color
  fonttitle=\bfseries,      % Title font style
  title=#1,                 % Title text argument
  enhanced,                 % Allows advanced skin features
  attach boxed title to top left={yshift=-2mm, xshift=2mm}, % Prompt-style title placement
  boxed title style={colback=purple!75!black},
  sharp corners=all,        % Slightly rounded modern corners
  drop shadow               % Adds a subtle drop shadow
}

\lstdefinestyle{pythonstyle}{
  language=Python,
  basicstyle=\ttfamily\footnotesize,
  keywordstyle=\color{purple!75!black}\bfseries,
  stringstyle=\color{teal!70!black},
  commentstyle=\color{gray}\itshape,
  numberstyle=\tiny\color{gray},
  showstringspaces=false,
  breaklines=true,
  tabsize=4,
}

\newtcblisting{pythonpromptbox}[1]{%
  colback=gray!5,
  colframe=purple!75!black,
  fonttitle=\bfseries,
  title=#1,
  enhanced,
  attach boxed title to top left={yshift=-2mm, xshift=2mm},
  boxed title style={colback=purple!75!black},
  sharp corners=all,
  drop shadow,
  breakable,               % allow the code block to span pages
  listing only,
  listing options={style=pythonstyle},
}

\title{\LARGE \bf
 Co$\mathbf{^2}$Gen: Collaboratively Generating Task Plans for Decentralized Multi-Robot and Human Collaboration
}

\author{Yi Loo$^{1}$, Siddhant Shrivastava$^{1}$, Minh Long Pham$^{1}$ and Malika Meghjani$^{1}$% <-this % stops a space
\thanks{$^{1}$ The authors are with Singapore University of Technology and Design (SUTD), Singapore.
        {\tt\small \{yi\_loo,siddhant\_shrivastava, minhlong\_pham\}@mymail.sutd.edu.sg , malika\_meghjani@sutd.edu.sg}}%
}

\begin{document}

\maketitle
\thispagestyle{empty}
\pagestyle{empty}

%%%%%%%%%%%%%%%%%%%%%%%%%%%%%%%%%%%%%%%%%%%%%%%%%%%%%%%%%%%%%%%%%%%%%%%%%%%%%%%%
\begin{abstract}

 Visual Language Action (VLA) models are designed to be generalist robot models that can be implemented across embodiments, tasks and environments. However, their ability to collaborate with other robots and humans remains underexplored. In this work, we propose a framework to automate the generation of multi-robot collaborative data for decentralized VLA training without requiring human teleoperation. Our framework first utilizes a decentralized collaborative planning approach to generate multi-robot collaborative robot plans in the form of an executable function. We then utilize the collaborative plan to produce collaborative demonstration data to finetune VLA models in a decentralized manner. We show that our approach produces demonstration data that when used to finetune VLA models on decentralized collaboration tasks matches the task performance of VLA models finetuned on human-teleoperated collaborative data.

% However it may be costly for humans to hand craft skill composition policies for new novel goals. We propose  a framework to leverage existing library of learned robot skills to perform novel multi-robot tasks though a collaborative planning process. Our proposed framework first utilizes inter-robot dialogue via LLMs to generate a joint execution policy via collaborative Monte Carlo Tree Search (MCTS) planning. The result is a Pythonic function that composes the individual robot skills into multi-robot collaborative execution plans. We validate our proposed approach on a multi-robot SO-101 arms setup on a suite of collaborative tasks and show that our proposed approach provides a cost effective approach the effectively adapts existing robot skill libraries into robust collaborative plans.

\end{abstract}

%%%%%%%%%%%%%%%%%%%%%%%%%%%%%%%%%%%%%%%%%%%%%%%%%%%%%%%%%%%%%%%%%%%%%%%%%%%%%%%%
\section{Introduction}
\label{sec:introduction}

Visual Language Action (VLA) models \citep{kimOpenVLAOpenSourceVisionLanguageAction2024a, intelligence$p_05$VisionLanguageActionModel2025, bjorck2025gr00t, fang2026molmoact2}  unify visual perception, language understanding and low-level robot actions into a single model, enabling them to generalize across embodiments, tasks and environments. Modern VLA models pre-train on large scale human demonstration datasets \citep{o2024open, khazatsky2024droid} and utilize Supervised Finetuning (SFT) on smaller task specific datasets for task adaptation \citep{kimOpenVLAOpenSourceVisionLanguageAction2024a, shukorSmolVLAVisionLanguageActionModel2025a}. If these VLA models are to be deployed on robots in real world environments, they invariably have to operate in the vicinity or even directly collaborate and coordinate with other robots and  humans. However, the modren VLA models' ability to perform collaborative tasks with other robots or even humans in zero-shot remains underexplored.  

In this paper, we show that VLA models, even when finetuned on task specific single robot demonstrations, fail to coordinate with each other on collaborative manipulation tasks as most of the pretrained data of VLA models come from non-collaborative single robot demonstration data. Hence, the VLA models fail to learn to coordinate their policy execution with one another. We then show that finetuning on demonstration data that is collected collaboratively in a multi-robot manner does improve VLA models performance on decentralized collaborative tasks. Furthermore, we extend these results to zero-shot human collaboration as well.

\begin{figure}[tp]
\centering
{\includegraphics[width=0.5\textwidth]{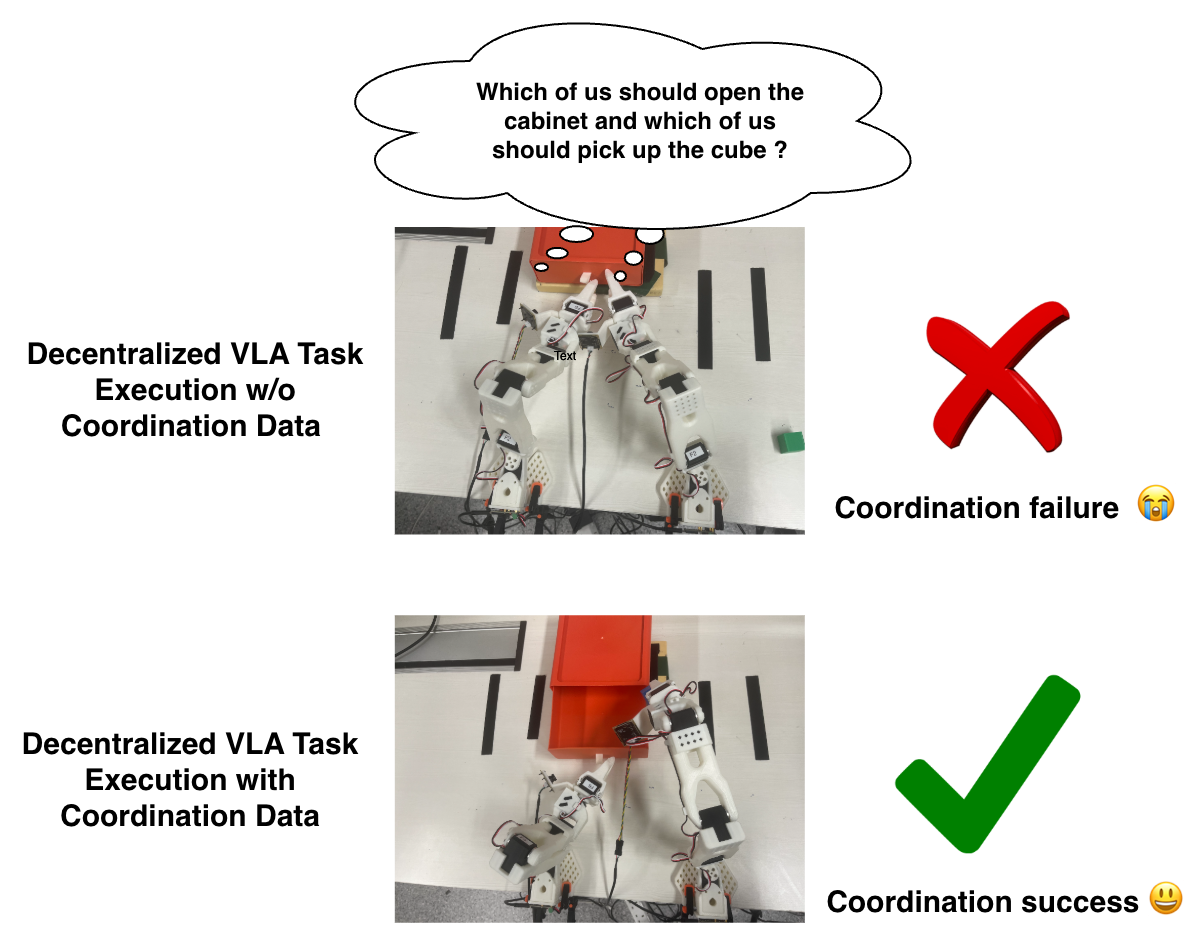}}

\caption{In decentralized task execution for a multi-robot setup, VLAs that are not trained on collaborative data fail to coordinate due to the lack of learned coordination behaviour.}
\label{fig:main_fig}
\vspace{-5pt}
\end{figure}

However, the data collection process for simultaneous multi-robot collaborative data can prove challenging due to either additional human cost for each additional robot or the lack of teleoperation interfaces to collect said demonstration data. To bridge this gap, we propose Co$^2$Gen: Collaborative Plan Generation for Decentralized VLA Training for Collaborative Tasks, a framework to autonomously generate collaborative data for multi-robot systems without multi-robot collaborative demonstrations. Our key insight is that collaborative robot demonstrations require the coordination and sequencing of each individual robot's sub-tasks to be conditioned on the successful sub-task execution and state of another robot. For example in Figure \ref{fig:main_fig}, the arm on the right should only perform its sub-task of placing its cube when the cabinet is in an open position. Hence, any collaborative plan would need to sequence robot sub-tasks conditioned on both environment state and sub-tasks of other robots.

Given a collaborative task prompt and environment state, our proposed approach, Co$^2$Gen, first leverages Visual Language Models (VLM) to generate a collaborative execution plan in the form of executable Pythonic function \citep{10161317, liang2023code}. We structure the plan generation process to be decentralized, where each robot utilizes a VLM to collaboratively propose and refine the generated execution plan using a collaborative Monte Carlo Tree Search (MCTS) process \citep{zu2025collaborative, amanickapraveen2026playing}. The resultant execution plans composes and calls on individual skills that are available for each robot and utilizes VLMs as a semantic verifier to check for sub-task preconditions for each robot skill execution.

We then utilize the generated collaborative execution plan to generate collaborative robot demonstration data together with a human verifier to check for task success. Finally, we finetune the VLAs on resultant demonstration data. Each robot's VLA model is finetuned separately in a decentralized manner from its local data stream. We show that it enables VLAs to match the task execution success rate of VLAs finetuned on human teleoperated collaborative demonstrations. We further show that they also retain the match the for zero-shot human-robot collaboration across manipulation tasks. To summarize, our contributions are:

\begin{enumerate}
    \item We show that VLA models fail to generalize to simple collaborative tasks even when finetuned on task specific single-robot demonstrations.
    \item We show that collaborative robot demonstrations finetuning  massively improves VLA models' performance on collaborative tasks.
    \item We propose Co$^2$Gen, an autonomous way to produce collaborative robot data through an executable Python function that composes single robot skills into collaborative plans. The collaborative plan is also generated in a decentralized manner by utilizing a collaborative MCTS process.
\end{enumerate}

\section{Related Work}
\label{sec:relatedwork}

\textbf{Foundation Models for Single and Multi-Robot Systems.} Since their introduction, many works leverage the sematic reasoning capabilities of FMs for robot planning and task execution. Most notably, SayCan \cite{saycan2022arxiv} matches language instructions with low-level robot skills through a pre-trained value function for skill selection in open ended tasks. ProgPrompt \cite{10161317} and CaP \cite{liang2023code} leverage the code writing capabilities of LLMs to generate robot policies in the form of code  given access to low-level skills, code libraries and environment specifications for single robot task execution. In our work, we extend this capability to generate robot plans for collaborative tasks.

In the multi-robot context, SMART-LLM \cite{kannan2024smart} utilizes a central LLM planner to perform task decomposition and allocation for a heterogeneous multi-robot system whereas LaMMA-P \cite{zhang2025lamma} utilizes the central LLM planner to generate PDDL problem descriptions for each robot and a PDDL planner at the task execution level. RoCo \cite{mandi2024roco} utilizes multi-agent dialog  to generate sub-task plain which then fed to a centralized motion planner. For our approach, we utilized multi-agent dialog to perform collaborative tree search to generate  a programmatic collaborative plan that decomposes and allocates sub tasks while also coordinating sub-task sequencing by checking for sub-task pre-conditions. Each robot then executes their sub-task via a decentralized planner.

\textbf{VLA models for Multi-Robot Tasks.} Though initial VLA models are primarily applied to single robot tasks \cite{brohan2022rt, kimOpenVLAOpenSourceVisionLanguageAction2024a, intelligence$p_05$VisionLanguageActionModel2025}, recent works have explored applying VLA models to multi-robot systems, primarily on the long standing problem of bi-manual manipulation \cite{koga1992experiments, lu2025anybimanual}. These approaches utilize information sharing between arms to coordinate each of their policies. InterAct \cite{pmlr-v270-lee25a} uses Hierarchical Attention Encoder to facilitate information sharing between robot arm policies. TwinVLA \citep{im2026twinvla} utilizes cross attention between the VLA models of both robot arms for information exchange and coordination. MolmoAct2 \citep{fang2026molmoact2} couples an embodied-reasoning VLM backbone with a flow-matching continuous action expert to enable coordinated dual-arm manipulators. SkillVLA \citep{zhai2026skillvla} proposes a hierarchical reasoning approach to bi-manual manipulation, where a high level VLM generates sub-tasks for each individual arm which are then passed on to low-level VLM linked to each arm's action experts. In our work, we extend bi-manual manipulation to a true multi-robot system where each arm has a separate policy via a VLA model. To our knowledge, we are the first to explore VLA models' ability to collaboratively execute tasks in a decentralized manner and to collaborate with humans in a zero shot manner. 
% Steerable VLA

% \cite{wang2026insight}
% \cite{wang2026insight}

% Multi-Robot with FM

% \cite{kannan2024smart}
% \cite{mandi2024roco}
% \cite{zhang2025lamma}
% \cite{wan2025embodiedagent}
% \cite{zu2025collaborative}
% \cite{zhao2023large}
% \cite{zhang2024building}

% FM for robot planning

% \citep{pmlr-v270-duan25a}
% \citep{huang2024rekep}
% \cite{hu2023look}
% \cite{athalye2025pixelspredicateslearningsymbolic}
% \cite{yang2025guiding}

% PDDL

% \cite{ye2026unidomain}

%===============================================================================

\section{Problem Formulation}
\label{sec:preliminaries}

% \subsection{Problem Formulation}
In this work, we focus on the multi-robot collaboration problem. Given a collaborative task, specified via natural language prompt, $\tau$, we would like to obtain a collaborative plan $\mathcal{P}$  that can generate a collaborative robot trajectory that corresponds with $\tau$. We would like the plan $\mathcal{P}$ to generate successful and varied robot episodes that can substitute for multi-robot demonstrations to fine-tune a VLA model. We assume the collaborative planner has access to a set of single robot skills $\mathcal{S} = \mathcal{S}_1 \bigcup \mathcal{S}_2 \bigcup \dots \bigcup \mathcal{S}_N$ where  $\mathcal{S}_n$ is the $n$-th robot's set of skills. Each robot may share common skills but also possess unique skills i.e. $\bigcap^{N}_{n=1} \mathcal{S}_n \neq \bigcup^{N}_{n=1} \mathcal{S}_n$. For this work we focus on a dual arm ($N=2$) multi-robot collaboration setup.

\subsection{Robot Setup}
\begin{figure}[htp]
\centering
{\includegraphics[width=0.5\textwidth]{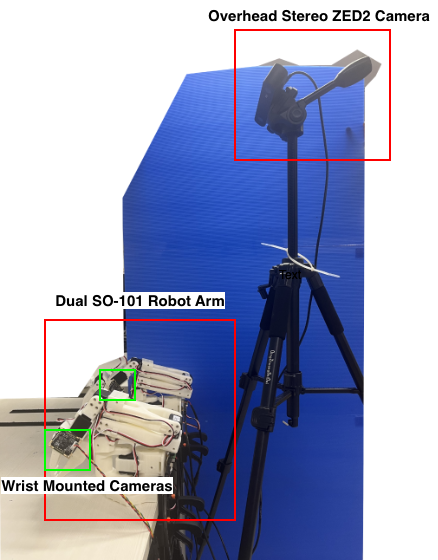}}

% \vspace{5cm}

\caption{Our bi-manual arm setup for all of collaborative tasks.}
\label{fig:setup}
% \vspace{-6mm}
\end{figure}

We conduct all of our experiments on a bi-manual low-cost SO-101 robot arm with an overhead stereo Zed2 Camera. Each robot also has a wrist camera attached near the gripper. We show our setup in Figure \ref{fig:setup}. 

% which we frame as an options learning framework. Each robot, possess a set of skills $\mathcal{I}_n$, Robots may share skills but also may also possess specialized skills, i.e. $\bigcap^{N}_{n=1} \mathcal{I}_n \neq \bigcup^{N}_{n=1} \mathcal{I}_n$. Given a joint task goal $G$, we are interested in constructing a joint policy via skill composition  

%===============================================================================

\section{Method}
\label{sec:method}

In this section we outline our proposed methodology. In Section \ref{Plan generation}, we outline our approach of generating functional collaborative robot plans in the form of a Pythonic function by utilizing a decentralized collaborative Monte Carlo Tree Search (MCTS) process. In Section \ref{Task execution}, we describe how the executable plan connects environmental perception with low-level robot actions to execute a collaborative task. In Section \ref{Data Generate}, we describe how we utilize the resultant robot plan  with a human verifier to collect collaborative robot data. Finally in Section \ref{Decen vla}, we describe how to train VLAs in a decentralized manner using the collaborative data for decentralized collaborative task execution and Human-Robot Collaboration.

\subsection{Collaborative Plan Generation Module.}
\label{Plan generation}

 \begin{figure*}[htp]
\centering
{\includegraphics[width=\textwidth]{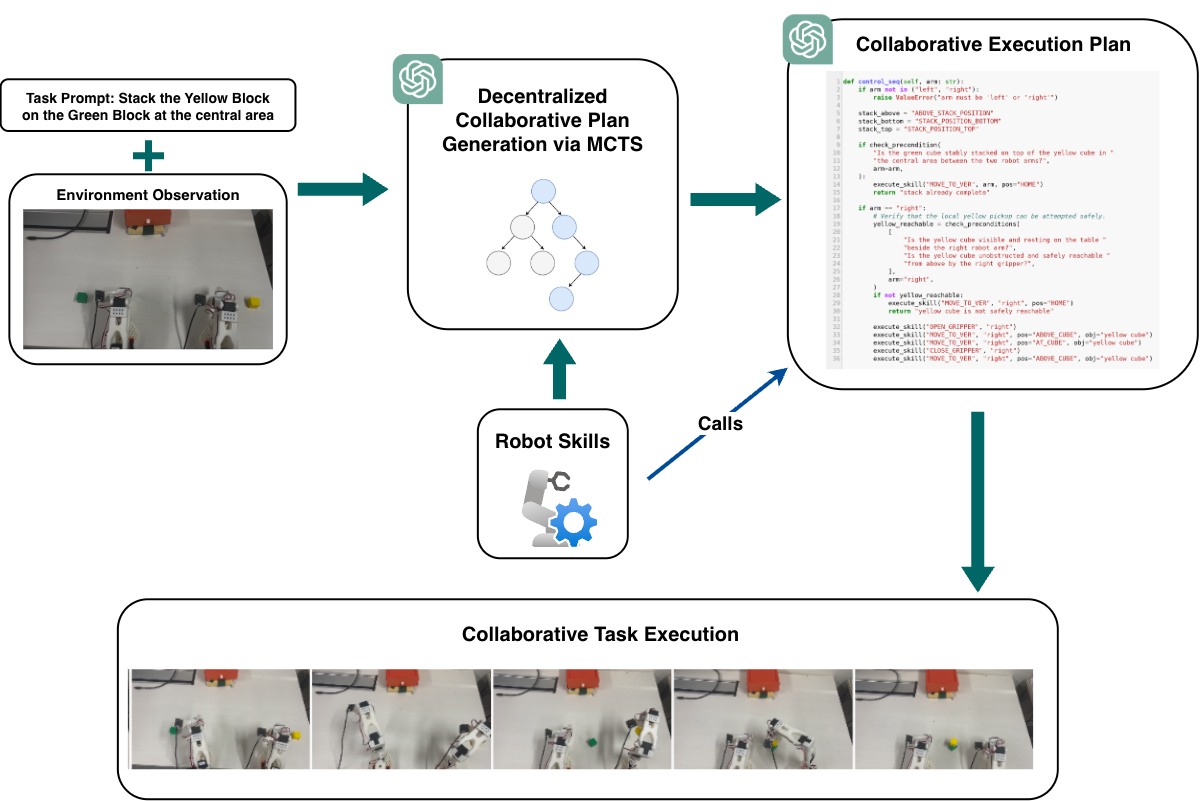}}

% \vspace{5cm}

\caption{The system diagram for our proposed collaborative plan generation approach.}
\label{fig:plan_generation_diag}
% \vspace{-6mm}
\end{figure*}

An effective multi-robot execution plan would need to reason about the constraints and preconditions to schedule and allocate  sub-tasks to the appropriate robot.  For example, an allocated sub-task or skill of one robot should only be scheduled to be executed after the successful execution sub-task execution from another robot. To that end, we identify three crucial features of such a plan: (i) Identify sub-task pre-conditions, (ii) Perform sub-task allocation and (iii) Schedule sub-tasks for overall task  coordination.  

We propose a collaborative plan generation module that leverages  VLMs to generate the collaborative plan in form of a executable Pythonic function. Our motivation for utilizing a code based execution plan is two-fold. First, LLMs and VLMs have demonstrated their ability to generate functions \citep{10161317, liang2023code} in robotic plan generation that connects environmental inputs with relevant robot skills. In the context of multi-robot collaboration, this allows the function to condition low-level robot skills on environment feedback. Second, a code-based robot execution plan enables decentralized parallel execution for multi-robot systems. 

We also design the plan generation module itself to be fully collaborative and decentralized. We adopt the approach by  \citep{ zu2025collaborative} to generate and iteratively refine the collaborative executable function via collaborative MCTS. Each of the two robot's plan generator is represented as a VLM agent, one assigned as an actor and the other a critic. Both actor and critic are given the overhead camera observation of the task environment together with a collaborative task prompt and the set of available single robot skills $\mathcal{S}$. The generation process starts with the actor agent proposing an initial execution function. The critic then evaluates and scores the generated function and provides targeted feedback to the actor. This initial proposal and critique serve as the root node $p_0$ of the tree-search process, which iteratively executes the following four steps to populate the tree:

\textbf{Selection} 
At each iteration, the planning module starts at the root and traverses the tree based on the UCT selection policy until a leaf node is reached. The UCT policy is defined as follows:

\begin{equation}
UCT(c_i) = V(c_i) + w\sqrt{ln \frac{N(p_i)}{N(c_i)}}
\end{equation}

where $N(c_i)$ is the number of visits for node $c_i$, $(V(c_i)$ is back-propagated value of  the subtree from $c_i$, $w$ is the exploration coefficient and $p_i$ is the parent of $c_i$.

\textbf{Expansion} 
Once a leaf node is reached, the module then expands the tree with $n$ new child nodes. Each child node's plan is a refinement of the parent node's plan based on the critique in the parent node.

\textbf{Evaluation} 
The critic agent gives a score $r$ for each generated plan based on three different criteria, the score of each criteria range from $1$ to $5$ for a maximum score of $15$. The three scoring criteria for each plan are: (i) Progress Score evaluates how much the generated plan makes meaningful progress towards the task. (ii) Distribution Score evaluates how much the generated plan allocates the appropriate sub-task to each robot. (iii) Condition Check Score evaluates whether the plan utilizes environment feedback to check for sub-task pre-conditions.

\textbf{Backpropagation}
The total critic score $r$ is back-propagated from the evaluated node to the root. The visitation count and value at each node along the traversal path are updated as follows:
\begin{equation}
N'(c_i) \leftarrow N(c_i) + 1, \quad \forall c_i \in \{c_0, \ldots, c_d\}
\end{equation}
\begin{equation}
V'(c_i) \leftarrow \frac{V(c_i)N(c_i) + r}{N'(c_i)}, \quad \forall c_i \in \{c_0, \ldots, c_d\}
\end{equation}
where $d$ is the current depth of the tree.

\textbf{Final Plan Selection}
The tree-search process terminates when the tree reaches a maximum depth $d_{max}$. A plan node is considered solved if the score is above a preset threshold score $r_{thres}$. The final execution plan is retrieved from the highest-value solved node in the tree. If no solved node exists, the node with the highest back-propagated value $V(c_i)$ is selected as a fallback. We show a diagram of our overall collaborative plan generation framework in Figure \ref{fig:plan_generation_diag} and an excerpt of the generated plan in Figure \ref{fig:example_plan}.

% \begin{figure}
%     \centering
%     \input{test_is.pdf_tex}
% \end{figure}

\begin{figure}[htp]
\centering
\begin{pythonpromptbox}{Generated Collaboration Plan}
# Alice waits until Bob has cleared the shared stacking workspace and the
# yellow cube is visibly established as the base.
yellow_ready = False
for _ in range(40):
  yellow_ready = check_precondition("Is the yellow cube placed in the central area between the two robot arms?", arm=arm,)
  if yellow_ready:
    break
  time.sleep(0.25)
    
if not yellow_ready:
  execute_skill("MOVE_TO_VER", arm, pos="HOME")
  return "Left arm did not observe the yellow stack base; aborted safely."
    
# Brief settling time keeps the left arm from entering the center while the
# right arm is completing its retreat.
time.sleep(0.75)
    
green_already_stacked = check_precondition("Is the green cube resting on top of the yellow cube?",arm=arm,)
    
if not green_already_stacked:
  execute_skill("MOVE_TO_VER", arm, pos="ABOVE_CUBE", obj="green cube")
    .
    .
    .
  execute_skill("MOVE_TO_VER", arm, pos="HOME")
  return "Left arm finished placing the green cube on the yellow cube."
\end{pythonpromptbox}
\caption{Collaborative plan generated by our MCTS procedure. The plan utilizes natural language prompts to check for pre-conditions before executing subsequent sub-tasks, allowing the robots to coordinate their actions during task execution.}
\vspace{-15pt}
\label{fig:example_plan}
\end{figure}

\subsection{Task Execution Module}
\label{Task execution}

We then utilize the generated collaborative execution plan to produce collaborative trajectories by connecting the semantic understanding of the environment with low-level robot skills. On the semantic understanding side, the collaborative plan uses VLMs to query pre-conditions in the environment via the overhead camera and individual wrist cameras. The low level robot skills consist of a perception module and trajectory planner. During task execution, the perception module utilizes hybrid blob detection and Yolov8 \citep{yolov8_ultralytics} via the overhead depth camera to obtain rough object coordinate estimates. The trajectory planner then utilizes an Inverse Kinematic (IK) solver \citep{duclusaud2025placo} to generate robot trajectories towards the detected object coordinates. Finally, the robot arms use their wrist cameras to get precise object coordinates for grasping.

%  \begin{figure}[H]
% \centering
% {\includegraphics[width=\textwidth]{}}

% \vspace{5cm}

  % execute_skill("OPEN_GRIPPER", arm)
  % execute_skill("MOVE_TO_VER", arm, pos="AT_CUBE", obj="green cube")
  % execute_skill("CLOSE_GRIPPER", arm)
  % execute_skill("MOVE_TO_VER", arm, pos="ABOVE_CUBE", obj="green cube")
  % execute_skill("MOVE_TO_VER", arm, pos="ABOVE_STACK_POSITION")
  % execute_skill("MOVE_TO_VER", arm, pos="STACK_POSITION_TOP", obj="yellow cube")
  % execute_skill("OPEN_GRIPPER", arm)
  % execute_skill("MOVE_TO_VER", arm, pos="ABOVE_STACK_POSITION")

% \caption{An example of the generated execution plan for the collaborative stacking task.}
% \label{fig:example_plan}
% \vspace{-6mm}
% \end{figure}

% Blob detection and Yolo hybrid with overhead depth camera to produce rough object coordinates, wrist camera refinement for precise object grasp coordinates 

% \citep{duclusaud2025placo} 

% \citep{yolov8_ultralytics}

\subsection{Collaborative Data Generation}
\label{Data Generate}

\begin{figure}[htp]
\centering
{\includegraphics[width=0.5\textwidth]{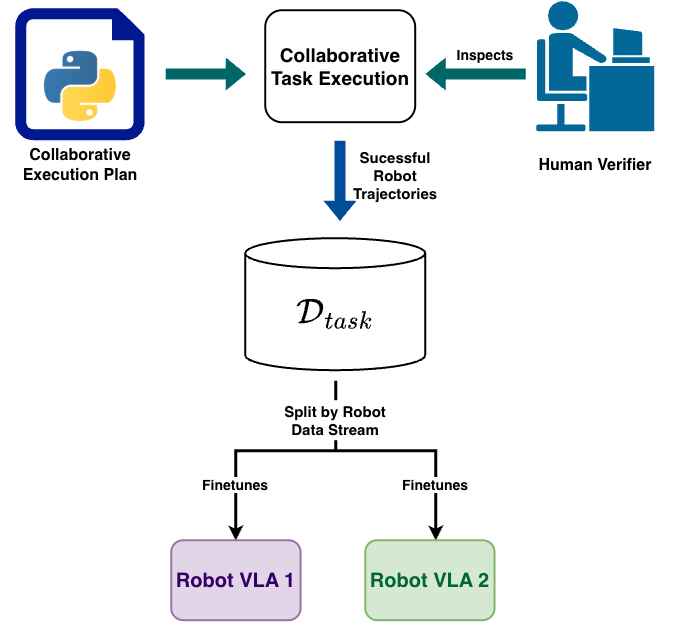}}

% \vspace{-10pt}

% \vspace{5cm}

\caption{Our framework for Collaborative Data Collection with a Human Verifier}
\label{fig:data_collection}
\vspace{-10pt}
\end{figure}

Here we describe our proposed framework for collaborative data collection. We incorporate a human into the data collection loop  as a task verifier. After each task execution, the human verifier checks if the task is successfully executed. Successfully executed robot episodes are then collected into a demonstration dataset $\mathcal{D}_{task}$. We chose a human verifier over direct VLM verification to mitigate any false positive trajectories from being included into the $\mathcal{D}_{task}$ due to VLM errors/hallucinations. We show our data collection framework diagrammatically in Figure \ref{fig:data_collection}.

\subsection{Decentralized VLA Finetuning}
\label{Decen vla}

Finally we perform Supervised Finetuning (SFT) on VLAs using the collected collaborative task dataset $\mathcal{D}_{task}$. We split $\mathcal{D}_{task}$ into robot specific data streams $\mathcal{D}^n_{task}$. Each data-stream consists of the robot's joint state, wrist camera stream and shares the overhead stereo camera stream. We then finetune each robot's VLA in a decentralized manner on their respective data streams.  We propose two distinct approaches to utilize $\mathcal{D}_{task}$ for VLA model training. The first approach trains a VLA model with $\mathcal{D}_{task}$ as is and second approach combines $\mathcal{D}_{task}$ with single robot human demonstration data. Single robot human demonstrations can produce diverse and varied trajectories but lack the coordination signal that is inherent to collaborative tasks, whereas the trajectories in $\mathcal{D}_{task}$ may lack the diversity of human demonstrations. Combining  $\mathcal{D}_{task}$,  with single human data, results in a hybrid dataset that covers both trajectory diversity and collaborative data without the need for multi-robot human teleoperation.

%===============================================================================

\section{Experiments}
\label{sec:experiments}

In this section, we evaluate Co$^2$Gen's ability to perform collaborative tasks in a decentralized manner. Chiefly, we would like to  answer the following research questions: 

\begin{enumerate}
    \item How does finetuning VLA models on collaborative demonstration data from both humans and Co$^2$Gen affect decentralized collaborative task performance?
    \item How does the resultant VLA perform on collaborative task with novel human partners in zero-shot?
\end{enumerate}

% For the Low level perception to control stack

\subsection{Evaluation Tasks}

\begin{figure}[h]
\centering
\vspace{-10pt}
{\includegraphics[width=0.4\textwidth]{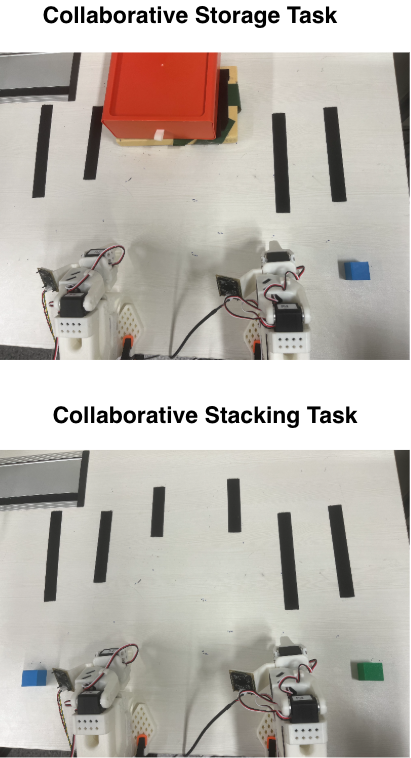}}

% \vspace{5cm}

\caption{The two different collaborative tasks we evaluate our method on. Each of the two tasks require both arms to coordinate their sub-task execution in order to achieve overall task success.}
\label{fig:tasks}
\vspace{-6mm}
\end{figure}

We evaluate our proposed approach on two  collaborative tasks. A \textbf{collaborative storage task} and \textbf{collaborative stacking task}. For the \textbf{collaborative storage task}, a single compartment pull-out cabinet is placed between the two arms. A colored block is placed near one of the arms. The task is to store the colored cube into the cabinet. For the \textbf{collaborative stacking task}, a single colored block is placed near each arm and the arms are tasked to stack the two blocks at the central region between the two arms. We show a visual representation of each of the two tasks in Figure \ref{fig:tasks}.

For each task, we denote the robot's collaborative role as the sub-tasks that are assigned to each arm. For the collaborative storage task, role $1$ is fulfilled by the arm that opens and closes the cabinet and role $2$ is the arm the places the cube in the cabinet. For the collaborative stacking task, role $1$ is fulfilled by the arm placing the bottom cube of the stack and role $2$ is the arm that stacks the top cube.

% - VLAs we finetune SmolVLA 
% \citep{shukorSmolVLAVisionLanguageActionModel2025a}

% \textbf{Evaluation Metrics}

% We evaluate the generated plans 

% \citep{shukorSmolVLAVisionLanguageActionModel2025a}

\subsection{Implementation Details}

We use GPT 5.6 \citep{openai2026gpt56}, specifically \texttt{gpt-5.6-sol} as the VLMs for Collaborative Plan Generation and for Semantic Precondition Checking during collaborative task execution. For the collaborative plan generation MCTS we set the max tree depth $d_{max} = 5$. For the VLA model, we use the lightweight SmolVLA \citep{shukorSmolVLAVisionLanguageActionModel2025a}, specifically \texttt{smolvla\_base} across all of our experiments. For each collaborative task we collect $50$ episodes of demonstration data and finetune for $20000$ steps. We also use DAgger \citep{ross2011reduction} to collect further correction data to improve the reliability of each model. The robot episodes are split evenly by collaborative roles for each arm. 

\begin{table}[htp]
    \centering
    \caption{Sub-Tasks for Each Collaborative Task. Arm $i$ denotes the Arm that takes on Role $i$ of the task. A Sub-task is considered a success if the robot arm executes it as described without any collision with the other arm.}
    \begin{tabular}{>{\raggedright\arraybackslash}p{0.8\linewidth}c}
        \toprule
        \multicolumn{2}{c}{Collaborative Stacking Task  (Max Score $=14$)} \\
        \midrule
        \multicolumn{1}{c}{Sub-Task} & Score \\
        \midrule
          Arm 1 successfully picks up cube from its vicinity & $2$  \\
          Arm 2 successfully picks up cube from its vicinity & $2$  \\
          Arm 1 successfully places up the bottom cube at the central region between the arms & $2$  \\
          Arm 1 moves clears of the central region after placing the bottom cube& $3$  \\
          Arm 2 successfully stacks the top on the bottom cube & $5$ \\
        \midrule
          \multicolumn{2}{c}{Collaborative Storage Task (Max Score $=14$)} \\
          \midrule
          \multicolumn{1}{c}{Sub-Task} & Score \\
          \midrule
          Arm 1 successfully grips on the cabinet handle & $3$  \\
          Arm 1 successfully opens the cabinet & $3$  \\
          Arm 2 successfully picks up cube from its vicinity & $1$  \\
          Arm 2 successfully places the cube in the open cabinet & $3$  \\
          Arm 2 moves clear of the cabinet after placing the cube & $2$  \\
          Arm 1 successfully closes the cabinet & $2$ \\
          % \midrule
          % \multicolumn{2}{c}{Dish Cleaning Task (Max Score $=13$)}\\
          % \midrule
          % \multicolumn{1}{c}{Sub-Task} & Score \\
          % \midrule
          % Arm 1 successfully grips onto and picks up the wooden plate & $5$  \\
          % Arm 1 successfully holds the wooden plate at the central area between the arms & $3$  \\
          % Arm 2 successfully pick up the cleaning sponge & $2$  \\
          % Arm 2 successfully cleans the wooden plate with cleaning sponge & $3$  \\

        \bottomrule
    \end{tabular}
    
    \label{tab:sub_tasks}
\end{table}

% \subsection{Evaluation of Collaborative Plan Execution}

% We first evaluate the effectiveness of of generated plans on the three collaborative tasks. We evaluate them across three different metrics. \textit{TS} represents Task Success rate, which is a binary check of whether the final objective of each task was executed successfully. Additionally we also report the \textit{PS}, which is the plan success rate, which is success rate of the generated plan independent from any physical grasping failures. For each task, we also introduce a quantitative measure of the success of each sub-task, specific to each of the three tasks at both the plan and task level. We specify the sub-tasks for each collaborative task in Table \ref{tab:sub_tasks}. We layout the For each metric, we average  across $5$ separately generated collaborative plans. 

% \textbf{Results:}

% \begin{table}[htp]
%     \centering
%     % \vspace{-7.5mm}

%     \caption{Results Collaborative Plan Execution}
%     % \footnotesize{
%      % \vspace{-2mm}
%     \begin{tabular}{c|cccc}
%         \toprule
%         &\textit{TS} &\textit{PS} & \textit{STS-E} & \textit{STS-P}\\
%         \midrule
%         Stacking Task      &  &&  \\   
%         Storage Task  &  &&  \\ 
        
%         Cleaning Task   &  &&  \\ 
%         Average    &  &&  \\ 
%         \bottomrule
%     \end{tabular}
%     \label{tab:plan_exec_results}
% \end{table}

% - We compare our method with Decentralized 
%===============================================================================
\subsection{Decentralized Collaborative Task Execution with VLAs}
\label{vla_eval}

We evaluate the VLA models finetuned on the demonstration data produced by Co$^2$Gen, which we  denote as $VLA_{Co^{2}Gen}$ across the two collaborative tasks and compare them against two other baselines. We present two different versions of $VLA_{Co^{2}Gen}$. The vanilla $VLA_{Co^{2}Gen}$ is trained on data solely collected from generated collaborative execution plan. $VLA_{Co^{2}Gen + SA}$ merges the data collected from Co$^2$Gen with the single arm data. For decentralized task execution evaluation we run a pair of finetuned VLA models across 20 trials, each VLA model independently executes its task in the shared environment. We also note that for decentralized task execution each policy does not share any information during training and can only coordinate via their policy actions.  We run the models until the task is successfully completed, the task has irrecoverably failed (i.e. objects are knocked beyond reach) or after a timeout. We detail the two baselines we compare against below:

\begin{itemize}

    \item \textbf{VLAs fine-tuned on task specific single-arm human demonstration data $\mathbf{VLA_{SA}}$}. We collect role-specific single arm data for each of the two tasks across both arms. Crucially the demonstrations do not contain any sub-task execution from the other robot arm.
    \item \textbf{VLAs fine-tuned on task specific collaborative bi-manual human demonstration data $\mathbf{VLA_{BM}}$} We collect full bi-manual human teleoperated collaborative  demonstrations and similar to $VLA_{Co^{2}Gen}$ we finetune both robot arm's VLA in a decentralized manner on the split data streams.
\end{itemize}

\begin{figure*}[htp]
    \centering
    \includegraphics[width=\textwidth]{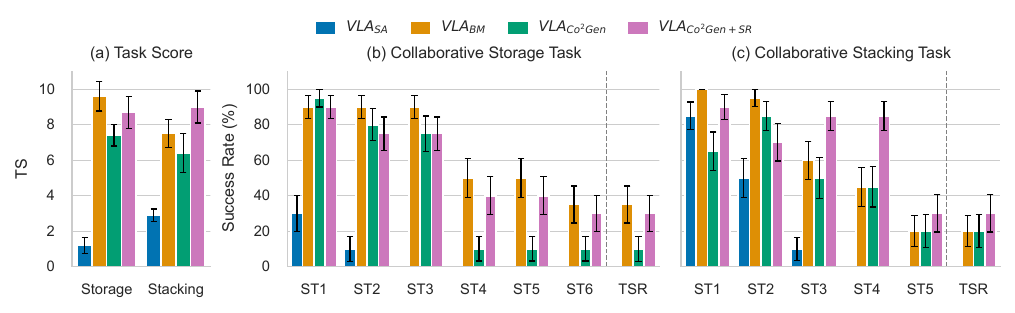}
    
    \caption{Results for the Collaborative Storage and Stacking Tasks across 20 task executions. (a) Overall task score (TS) as scored in Table \ref{tab:sub_tasks}. (b), (c) Sub-Task success rates (ST$x$ SR) and overall task success rate (TSR) for the Storage and Stacking Tasks respectively. Error bars denote the standard error.}
    \label{fig:task_results}
\end{figure*}

For evaluation we report  the overall decentralized collaborative task success rate (TSR) as well as task  score (TS) which quantifies the sub-tasks that were successfully executed the list of sub-tasks as well as their respective score are outlined in Table \ref{tab:sub_tasks}. We report our results for decentralized task execution across both tasks in Fig. \ref{fig:task_results}. We state the three major findings from these results as follows:

% \begin{figure*}[htp]
% \centering
% % {\includegraphics[width=\textwidth]{}}

% \vspace{5cm}

% \caption{Qualitative results across 3 tasks}
% \label{fig:qualitative_results}
% % \vspace{-6mm}
% \end{figure*}

% We report our results for decentralized task execution across both tasks in Fig. \ref{fig:task_results}. We state the three major findings from these results as follows:

\textbf{Finding 1: Single Robot task demonstrations fail for collaborative tasks.} Perhaps unsurprisingly, $VLA_{SA}$ has the lowest task performance across both tasks. The lack of multi-robot collaborative presence leads to a resultant VLA policy that does not possess or recognize any coordination signals from its partner arm and fails to coordinate their sub-tasks. For the collaborative storage task, both arms decide to simultaneously reach for the cabinet even though one of the arms is supposed to carry the cube. For the stacking task, both arms consistently move towards a collision at the center due to them not learning any way to coordinate their stacking sequence.

\textbf{Finding 2: Multi-Robot task demonstrations improve collaborative task performance.} The VLA Models finetuned with bi-manual data shows a significant improvement over $VLA_{SA}$ in both collaborative tasks. However, decentralized collaboration still proves to be challenging as the task success rate sits below 50 \% across both tasks. While $VLA_{BM}$ policies do coordinate their actions for collaboration, e.g. waiting for the other arm to stack the bottom cube or waiting for cabinet door to open before storing it, these coordination signals do not trigger consistently during task execution. We hypothesize that this is due to the VLA models being unable to capture the full coordination signals within the bi-manual collaboration data  due to the lack of information exchange during training. 

\textbf{Finding 3: Co$^2$Gen Data can bridge the gap between Single and Bi-manual VLA performances.} Our proposed approach of collecting robot demonstration data via generated robot plans result in marked improvement over $VLA_{SA}$. However the resultant VLA fails to complete sub-tasks well, as evidenced by $VLA_{Co^{2}Gen}$'s drop-off in the sub-task success rate for the Collaborative Storage Task. However when Co$^2$Gen data is combined with single arm demonstration data, it matches the task performance of $VLA_{BM}$ for storage task within the margin of error and exceeds its performance for the stacking task. The demonstration data from Co$^2$Gen adds the crucial coordination signal to the single arm demonstration data.  Furthermore as coordination in Co$^2$Gen is gated by condition checks,  the resultant data  sequences sub-task execution and coordination in a  structured manner as compared to human teleoperated collaboration data, which may have more ``blurry" boundary between sub-task executions. We hypothesize that this clearer separation in sub-tasks in Co$^2$Gen allow VLA models to more effectively learn coordination signals for decentralized collaborative task execution.

% allows VLAs to learn these cue while avoiding any risky skill executions that a tightly bounded skill and coordination cue will have.

\subsection{Human-Robot Collaboration}

We evaluate the VLA baselines in Subsection \ref{vla_eval} together with $VLA_{Co^{2}Gen}$ and $VLA_{Co^{2}Gen +  SA}$  on their ability to collaborate with human partners in zero-shot. In this setting, the human partner collaborates with autonomous an robot via teleoperation. The human partner is instructed to execute the collaboration task as per the task prompt.  Results are averaged over $10$ execution episodes across both  collaborative roles. We report the taskd success rate as well as the role specific task scores, i.e. sub task that are executed by the robot arm.

\begin{figure*}[htb]
    \centering
    \includegraphics[width=\textwidth]{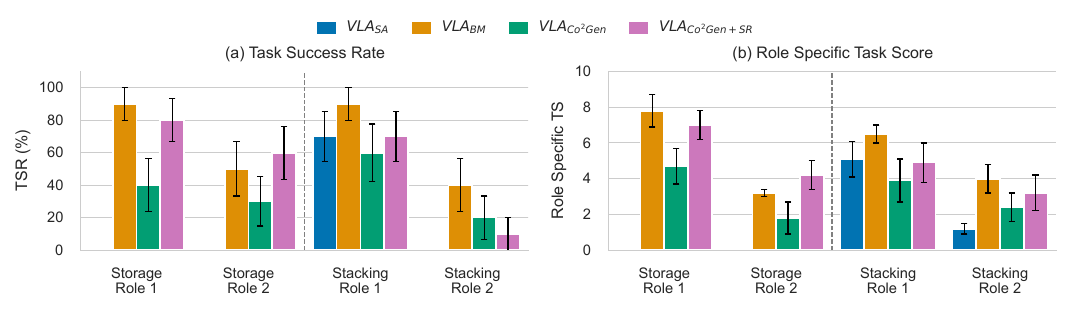}
    \caption{Results for Human-Robot Collaboration on the Collaborative Storage and Stacking Tasks for each role taken by the robot arm. (a) Task success rate (TSR). (b) Role specific task score (TS).}
    \label{fig:hrc_results}
\end{figure*}

\textbf{Results:} We show the Human-Robot Collaboration results for the storage and stacking tasks in Fig. \ref{fig:hrc_results}. Overall, task success rate skews higher than decentralized collaborative task execution in the previous subsection. We hypothesize that this is due to human partners being comparatively better at detecting and responding to coordination signals. Hence, the human partner can adapt its behaviour towards the VLA policies, which allow them to coordinate even with $VLA_{SA}$ performing the stacking task as role 1. However, we still show that the presence of collaborative data matters for Human-Robot Collaboration, as evidenced by the low performance of $VLA_{SA}$ in role 2 of the stacking task as well as both roles of the storage task.

\section{Conclusion}

In this work, we introduce Co$^2$Gen, a framework to generate data for VLAs to collaborate in a decentralized manner without the need for human teleoperated collaborative data. We show that a collaborative Python function generated via a collaborative MCTS process can act as a collaborative multi-robot execution plan by verifying sub-task preconditions through VLMs. We then show that the generated collaborative plan can substitute human teleoperation to produce collaborative multi-robot data for VLA finetuning. Finally, we show the VLA models finetuned on Co$^2$Gen data match the task performance of VLA models finetuned on human teleoperated collaborative data.

For future work, we hope to extend this work across different embodiments such as humanoids and mobile robots on longer horizon collaborative tasks in complex environments and in greater team sizes.

\label{sec:conclusion}

%===============================================================================

% \clearpage

%%%%%%%%%%%%%%%%%%%%%%%%%%%%%%%%%%%%%%%%%%%%%%%%%%%%%%%%%%%%%%%%%%%%%%%%%%%%%%%%

\section*{Acknowledgment}

This work was supported by TL@SUTD Seed Research Programme, which encourages innovative research in data and information exploitation.

\bibliographystyle{IEEEtranS}

\bibliography{ref}

\end{document}